\documentclass{article} 
\usepackage{iclr2021_conference,times}

\usepackage{amsmath,amsfonts,bm}

\def\eqref#1{equation~\ref{#1}}

\def\1{\bm{1}}

\DeclareMathAlphabet{\mathsfit}{\encodingdefault}{\sfdefault}{m}{sl}
\SetMathAlphabet{\mathsfit}{bold}{\encodingdefault}{\sfdefault}{bx}{n}

\usepackage{hyperref}
\usepackage{url}
\usepackage{graphicx}

\newcommand\blfootnote[1]{%
  \begingroup
  \renewcommand\thefootnote{}\footnote{#1}%
  \addtocounter{footnote}{-1}%
  \endgroup
}

\title{Targeted VAE: Variational and Targeted Learning for Causal Inference}

\author{Matthew J. Vowels, Necati Cihan Camgoz \& Richard Bowden  \\
Centre for Computer Vision, Speech, and Signal Processing\\
University of Surrey\\
Guildford, Surrey, UK \\
\texttt{\{m.j.vowels,n.camgoz,r.bowden\}@surrey.ac.uk} \\
}

\newcommand{\bftab}{\fontseries{b}\selectfont}

\newcommand{\indep}{\perp \!\!\! \perp}

\iclrfinalcopy 
\begin{document}

\maketitle

\begin{abstract}
Undertaking causal inference with observational data is incredibly useful across a wide range of tasks including the development of medical treatments, advertisements and marketing, and policy making. There are two significant challenges associated with undertaking causal inference using observational data: treatment assignment heterogeneity (\textit{i.e.}, differences between the treated and untreated groups), and an absence of counterfactual data (\textit{i.e.}, not knowing what would have happened if an individual who did get treatment, were instead to have not been treated). We address these two challenges by combining structured inference and targeted learning. In terms of structure, we factorize the joint distribution into risk, confounding, instrumental, and miscellaneous factors, and in terms of targeted learning, we apply a regularizer derived from the influence curve in order to reduce residual bias. An ablation study is undertaken, and an evaluation on benchmark datasets demonstrates that TVAE has competitive and state of the art performance.
\end{abstract}

\blfootnote{Accepted to Causal Learning (Special Track) at 2021 IEEE International Conference on Smart Data Services (SMDS).}

\section{Introduction}
The estimation of the causal effects of interventions or treatments on outcomes is of the upmost importance across a range of decision making processes, such as policy making \citep{Kreif2019}, advertisement \citep{Bottou2013}, the development of medical treatments \citep{Petersen2017}, and the evaluation of evidence within legal frameworks \citep{Pearl2009, Siegerink2017}. Despite the common preference for Randomized Controlled Trial (RCT) data over observational data, this preference is not always justified. Besides the lower cost and fewer ethical concerns, observational data may provide a number of statistical advantages including greater statistical power and increased generalizability \citep{Deaton2018}. However, there are two main challenges when dealing with observational data. Firstly, the group that receives treatment is usually not equivalent to the group that does not (treatment assignment heterogeneity), resulting in selection bias and confounding due to associated covariates. For example, young people may prefer surgery, older people may prefer medication. Secondly, we are unable to directly estimate the causal effect of treatment, because only the factual outcome for a given treatment assignment is available. In other words, we do not have the counterfactual associated with the outcome for a different treatment assignment to that which was given. Treatment effect inference with observational data is concerned with finding ways to estimate the causal effect by considering the expected differences between factual and counterfactual outcomes.

We seek to address the two challenges by proposing a method that enables the estimation of causal effects from observational data by leveraging techniques from the targeted learning literature. Specifically, Targeted Maximum Likelihood Estimation (TMLE) yields asymptotically efficient, unbiased, and doubly robust estimation of the causal effect, making it attractive as a causal inference method in its own right \citep{vanderLaan2011,vanderLaan2018, Schuler2016, vanderLaan2014}. We incorporate targeted learning into a variational latent model, trained according to the approximate maximum likelihood paradigm. Doing so enables us to infer hidden confounders from proxy variables in the dataset, and to estimate average treatment effects, as well as conditional treatment effects. Estimating the latter is especially important for treatments that interact with patient attributes, whilst also being crucial for facilitating individualized treatment assignment. Thus, we propose the Targeted Variational AutoEncoder (TVAE), undertake an ablation study and also compare our method's performance against alternatives on two benchmark datasets.\footnote{Code is available at  \url{https://github.com/matthewvowels1/TVAE_release}}

\begin{figure}[t!]
\centering
\includegraphics[width=0.8\linewidth]{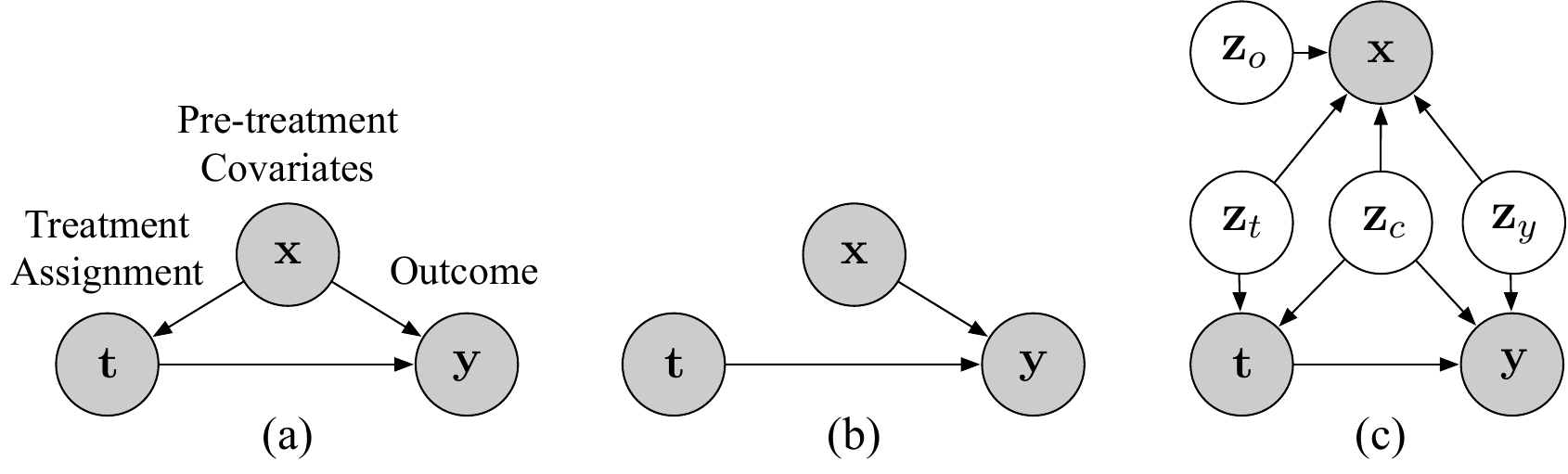}
\caption{Directed Acyclic Graphs (DAGs) for (a) the problem of estimating the effect of treatment $t$ on outcome $y$ with confounding $\mathbf{x}$. DAG (b) reflects an RCT. DAG (c) illustrates TVAE and is an extension of the DAG in \citep{Zhang2020}, where the structure is \textit{a priori} assumed to factorize into into risk $\mathbf{z}_y$, instrumental $\mathbf{z}_t$, and confounding factors $\mathbf{z}_c$. We extend their model with $\mathbf{z}_o$ to account for that fact that not all covariates will be related to treatment and/or outcome.}
\label{fig:PGMs}

\end{figure}

\section{Background}

\textbf{Problem Formulation:}~ 
A characterization of the problem of causal inference with no unobserved confounders is depicted in the Directed Acyclic Graphs (DAGs) shown in Figs.~\ref{fig:PGMs}(a) and \ref{fig:PGMs}(b). For an accessible overview of the relevant background concerning causal inference and graphs, consider \citep{Guo2020, Pearl2016, Vowels2021DAGS}. Fig.~\ref{fig:PGMs}(a) is characteristic of observational data, where the assignment of treatment is related to the covariates. Fig.~\ref{fig:PGMs}(b) is characteristic of the ideal RCT, where the treatment is unrelated to the covariates. Here, $\mathbf{x}_i \sim p(\mathbf{x}) \in \mathbb{R}^{m}$ represents the $m$-dimensional, pre-treatment covariates for individual $i$ assigned factual treatment $t_i \sim p(t|\mathbf{x})$ resulting in outcome $y_{i} \sim p(y|\mathbf{x},t)$. Together, these constitute dataset $\mathcal{D} = \{ [y_{i},  t_i, \mathbf{x}_i ] \}_{i=1}^N$ where $N$ is the sample size. We occasionally refer to a \textit{potential outcome} \citep{Imbens2015} for a particular treatment as $y(t)$, where the value of $t$ may or may not correspond to that which has been observed for a particular individual (\textit{i.e.}, it is a counterfactual outcome).

The conditional average treatment effect for an individual with covariates $\mathbf{x}_i$ may be defined as $\tau_i(\mathbf{x}_i) = \mathbb{E}[y_i|\mathbf{x}_i, \mbox{do}(t=1) - y_i|\mathbf{x}_i, \mbox{do}(t=0)]$, where the expectation accounts for the non-determinism of the outcome \citep{Jesson2020}. Alternatively, by comparing the post-intervention distributions when we intervene on treatment $t$, the Average Treatment Effect (ATE) is $\tau(\mathbf{x}) = \mathbb{E}_{\mathbf{x}}[\mathbb{E}[y|\mathbf{x}, \mbox{do}(t=1)] -\mathbb{E}[y|\mathbf{x}, \mbox{do}(t=0)]]$. Here, $\mbox{do}(t=t')$ indicates the intervention on $t$, setting it to a prescribed static value, dynamic value, or distribution and therefore removing any dependencies it originally had \citep{Pearl2009, vanderLaan2018, vanderLaan2011}. This scenario corresponds with Fig.~\ref{fig:PGMs}(b), where treatment $t$ is no longer a function of the covariates $\mathbf{x}$. 

Using an estimator\footnote{We use circumflex to designate an estimated (rather than true population) quantity.} for the conditional mean $Q(t,\mathbf{x})=\mathbb{E}(y|t,\mathbf{x})$, we can calculate the Average Treatment Effect (ATE) and the empirical error for estimation of the ATE (eATE).\footnote{For a binary outcome variable $y \in \{0,1\}$, $\mathbb{E}(y|t,\mathbf{x})$ is the same as $p(y|t,\mathbf{x})$.} The estimated Average Treatment Effect (ATE) and error on the estimation of ATE (eATE) are given in Eq.~\ref{eq:ATEstuff}. 

\begin{equation}
\begin{split}
\hat{\boldsymbol{\tau}}(\hat Q;\mathbf{x}) = \frac{1}{N} \sum_{i=1}^N (\hat Q(1,\mathbf{x}_i) - \hat Q(0,\mathbf{x}_i)), \\  \epsilon_{ATE} = |\tau(\mathbf{x}) - \frac{1}{N}\sum_{i=1}^N \hat{\boldsymbol{\tau}}(\hat Q;\mathbf{x}_i) |
\end{split}
\label{eq:ATEstuff}
\end{equation}

\begin{figure}[t!]
\centering
\includegraphics[width=0.7\linewidth]{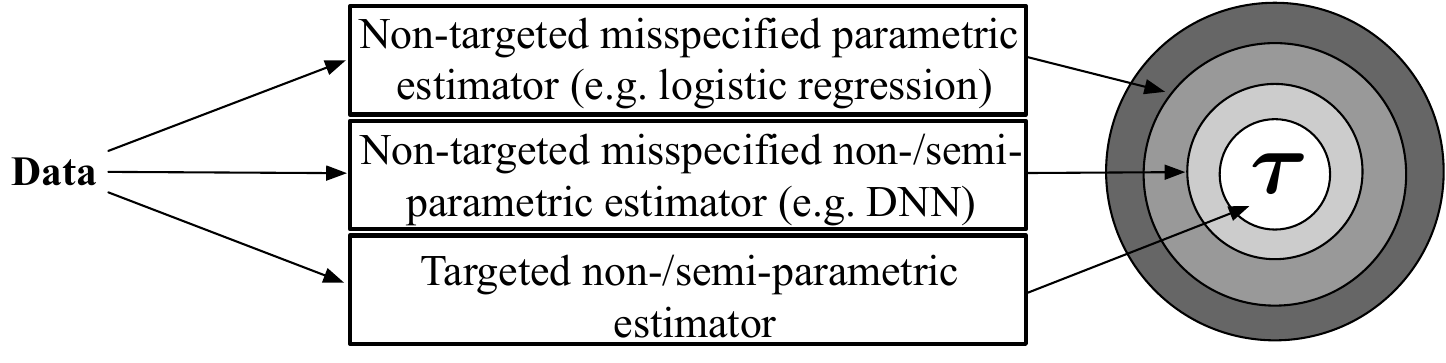}
\caption{Different methods for estimating the causal parameter $\boldsymbol{\tau}$ yield different levels of bias. Adapted from \citep{vanderLaan2011}.}
\label{fig:approaches}
\end{figure}

In order to estimate eATE we assume access to the ground truth treatment effect parameter $\boldsymbol{\tau}$, which is only possible with synthetic or semi-synthetic datasets. The Conditional Average Treatment Effect (CATE) may also be calculated on a per-individual basis and the Precision in Estimating Heterogeneous Effect (PEHE) is one way to evaluate a model's efficacy in estimating this quantity:

\begin{equation}
    \epsilon_{PEHE} =  \sqrt{ \frac{1}{N}\sum_{i=1}^N (\hat{\boldsymbol{\tau}}(\hat Q;\mathbf{x}_i) - \tau(\mathbf{x}_i))^2}
    \label{eq:PEHE}
\end{equation}
\textbf{The Naive Approach:}~
The DAG in Fig.~\ref{fig:PGMs}(a) highlights the problem with taking a naive approach to modeling the joint distribution $p(y,t,\mathbf{x})$. The structural relationship $t \leftarrow \mathbf{x} \rightarrow y$ indicates both that the assignment of treatment $t$ is dependent on the covariates $\mathbf{x}$, and that a backdoor path  exists through $\mathbf{x}$ to $y$. In addition to our previous assumptions, if we also assume linearity, adjusting for this backdoor path is a simple matter of adjusting for $\mathbf{x}$ by including it in a logistic regression. The naive method is an example of the uppermost methods depicted in Fig.~\ref{fig:approaches}, and leads to the largest bias. The problem with the approach is (a) that the graph is likely misspecified such that the true relationships between covariates as well as the relationships between covariates and the outcome may be more complex. There is also problem (b), that linearity is not sufficient to `let the data speak' \citep{vanderLaan2011, Vowels2021} or to avoid biased parameter estimates (\textit{i.e.}, functional misspecification). Using powerful nonparametric models (\textit{e.g.}, neural networks) may solve the limitations associated with linearity and interactions to yield a consistent estimator for $p(y|\mathbf{X})$, and such a model is an example of the middlemost methods depicted in Fig.~\ref{fig:approaches}. However, this estimator is not targeted to the estimation of the causal effect parameter $\boldsymbol{\tau}$, only predicting the outcome, and we require a means to reduce residual bias, such as targeted learning \citep{vanderLaan2011,vanderLaan2018,Schuler2016}. 

\textbf{Targeted Learning:} Targeted Maximum Likelihood Estimation (TMLE) \citep{Schuler2016, vanderLaan2011, vanderLaan2018} involves three main steps: (1) estimation of the conditional mean $\mathbb{E}(y|t,\mathbf{x})$ with estimator $\hat{Q}^0(t,\mathbf{x})$, (2) estimation of the propensity scores with estimator $\hat g(t|\mathbf{x})$, and (3) updating the conditional mean estimator $\hat Q^0$ to get $\hat Q^*$ using the propensity scores to attain an estimate for the causal parameter $\boldsymbol{\mathbf{\tau}}$. 

The propensity score for individual $i$ is defined as the conditional probability of being assigned treatment $g(t_i,\mathbf{x}_i) = p(t=t_i | \mathbf{x}=\mathbf{x}_i), \: \in [0,1]$ \citep{Rosenbaum1983}. The scores can be used to compensate for the relationship between the covariates and the treatment assignment using Inverse Probability of Treatment Weights (IPTWs), reweighting each sample according to its propensity score. Step (3) is undertaken using `clever covariates' which are similar to the IPTWs. They form an additional covariate variable ${H(1,\mathbf{x}_i) = g(1| \mathbf{x}_i)^{-1}}$ for individual $i$ assigned treatment, and $H(0,\mathbf{x}_i) = -g(1| \mathbf{x}_i)^{-1}$ for individual $i$ not assigned treatment. The notation when conditioning on a single numeric value implies an intervention (\textit{i.e.}, $g(1| \mathbf{x}_i) \equiv g(\mbox{do}(t=1)| \mathbf{x}_i)$). A logistic regression is then undertaken: $y = \sigma^{-1}[\hat Q^0(t,\mathbf{x})] + \epsilon \hat H(t,\mathbf{x})$ where $\sigma^{-1}$ is the logit/inverse sigmoid function, $\hat Q^0(t,\mathbf{x})$ is set as a constant, suppressed offset (\textit{i.e.}, no associated regression coefficient) and $\epsilon$ represents a fluctuation parameter which is to be estimated from the regression. Once $\epsilon$ has been estimated, we acquire an updated estimator:
\begin{equation}
\begin{split}
\hat Q^1(\mbox{do}(t=t'),\mathbf{x}) = \\ \sigma \left[ \sigma^{-1}[\hat Q^0(t,\mathbf{x})] + \hat \epsilon \hat H(t, \mathbf{x}) \right]
\end{split}
\label{eq:update}
\end{equation}
Here $t'$ is the particular interventional value for $t$. This equation tells us that our new estimator $\hat Q^1$ is equal to the old estimator balanced by the corrective $\epsilon H(t, \mathbf{x})$ term. This term adjusts for the bias associated with the propensity scores. When the $\epsilon$ parameter is zero, it means that there is no longer any influence from the `clever covariates' $H()$. The updated estimator $\hat Q^1$ can then be plugged into the estimator for $\hat{\boldsymbol{\tau}}(\hat Q^1;\mathbf{x})$. When the optimal solution is reached (\textit{i.e.}, when $\epsilon = 0$), the estimator $\hat Q^*$ also satisfies what is known as the efficient Influence Curve (IC), or canonical gradient equation \citep{Hampel1974, vanderLaan2011, Kennedy2016}: 
\begin{equation}
\begin{split}
0 = \sum_{i=1}^N IC^*(y_i, t_i,\mathbf{x}_i) = \\\sum_{i=1}^N \left[ \hat H(t_i, \mathbf{x}_i)(y_i - \hat Q(t_i, \mathbf{x}_i)) + \right.\\ \left. \hat Q(1, \mathbf{x}_i) - \hat Q(0, \mathbf{x}_i) - \boldsymbol{\tau}(Q;\mathbf{x}) \right]
\end{split}
\label{eq:EIC}
\end{equation}
where $IC(y_i, t_i,\mathbf{x}_i)$ represents the IC, and $IC^*(y_i, t_i,\mathbf{x}_i)$ represents the efficient IC for consistent $\hat Q$ and $\hat g$. It can be seen from the right hand side Eq.~\ref{eq:EIC} that at convergence, the estimator and its corresponding estimand are equal: $y_i = \hat Q(t_i, \mathbf{x}_i)$ and $\hat Q(1, \mathbf{x}_i) - \hat Q(0, \mathbf{x}_i) =  \boldsymbol{\tau}(Q;\mathbf{x})$. Over the whole dataset, all terms in Eq.~\ref{eq:EIC} `cancel' resulting in the mean $\bar{IC}=0$. As such, the logistic regression in Eq.~\ref{eq:update} represents a solution to the IC via a parametric submodel. 

The TMLE method provides a doubly robust, asymptotically efficient estimate of the causal or `target' parameter, and these theoretical guarantees make it attractive for adaptation into neural networks for causal effect estimation.

\section{Methodology}
\begin{figure*}[t!]
\centering
\includegraphics[width=1\linewidth]{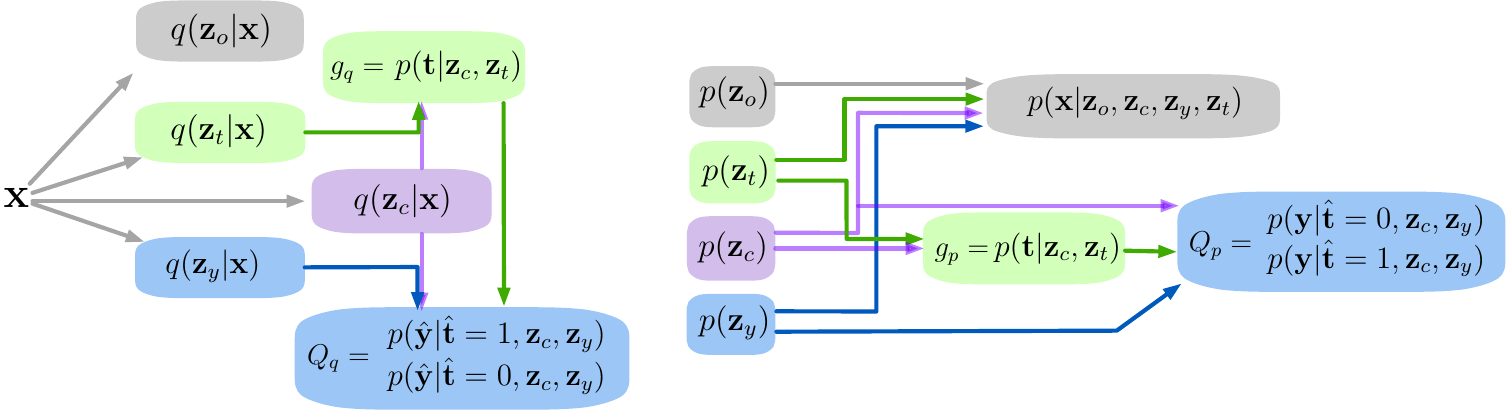}
\caption{The block-diagram for Targeted VAE. Dashed boxes indicate the variationally inferred latent variables $\mathbf{z}_o$, $\mathbf{z}_c$, $\mathbf{z}_t$, and $\mathbf{z}_y$. Arrows indicate functions, and colors distinguish treatment (green), outcome (blue), covariates (grey), and confounder (purple) related entities.}
\label{fig:TVAE}
\end{figure*}

In this section we present the Targeted Variational  AutoEncoder (TVAE), a deep generative latent variable model that enables estimation of the average and conditional average treatment effects (ATE and CATE resp.) via the combination of amortized variational inference techniques and Targeted Maximum Likelihood Estimation (TMLE). A top-level diagram for TVAE is shown in Fig.~\ref{fig:TVAE} and follows the structure implied by the DAG in Fig.~\ref{fig:PGMs}(c).

\textbf{Assumptions:} As is common \citep{Yao2020, Guo2019, Rubin2005, Imbens2015} when undertaking causal inference with observational data, we make a number of assumptions: (1) Stable Unit Treatment Value Assumption (SUTVA): the potential outcomes for each individual or data unit are independent of the treatments assigned to all other individuals, such that there are no interactions between individuals. (2) Positivity: the assignment of treatment probabilities are all non-zero and non-deterministic $p(t=t_i | \mathbf{x}=\mathbf{x}_i) > 0, \: \forall \; t \mbox{ and } \mathbf{x}$. (3) That all confounders have been inferred via noisy proxies present in the observed data \citep{Montgomery2000, Louizos2017b}, such that the likelihood of treatment for two individuals with the same inferred latent covariates is equal, and the potential outcomes for two individuals with the same latent covariates are also equal $s.t. \: y(1),y(0) \indep t|\mathbf{z}$  and   $t \indep  (y(1),y(0))|\mathbf{z}$, which constitutes a form of conditional exchangeability. In relation to assumption (3), see discussion below on identifiability.

\textbf{TVAE:} If one had knowledge of the true causal DAG underlying a set of data, one could undertake causal inference without being concerned for issues relating to structural misspecification. Unfortunately, and this is particularly the case with observational data, we rarely have access to this knowledge. Quite often an observed set of covariates $\mathbf{x}$ are modelled as a group of confounding variables (as per the DAG in Figure \ref{fig:PGMs}a). Furthermore, and as noted by \citep{Zhang2020}, researchers may in general be encouraged to incorporate as many covariates into their model as possible, in an attempt to reduce the severity of the ignorability assumption. However, including more covariates than is necessary leads to other problems relating to the curse of dimensionality and (in)efficiency of estimation. 

A large set of covariates may be separable into subsets of factors such as instrumental, risk, and confounding factors. Doing so helps us to match our model more closely to the true data generating process, as well as to improve estimation efficiency by `distilling' our covariate adjustment set. Prior work has explored the potential to discover the relevant confounding covariates via Bayesian networks \citep{Haggstrom2017}, regularized regression \citep{Kuang2017}, and deep latent variable models based on Variational Autoencoders (VAEs) \citep{Zhang2020, Louizos2017b, Wu2021}. The first two methods \textit{identify} variables (and are variable selection algorithms), whereas VAEs \textit{infer} them, and learn compact, disentangled representations of the observations. The benefit of the latter approach is that it (a) infers latent variables on a datapoint-by-datapoint basis (rather than deriving subsets from population aggregates), (b) under additional assumptions, VAEs have been shown to infer hidden confounders in the presence of noisy proxy variables, thereby potentially reducing the reliance on ignorability \citep{Louizos2017b, Lowe2020ACD}, and (c) makes no assumptions about the functional form used to map between covariate and latent space. 

\textbf{Variational Inference:} In general terms, variational inference is concerned with maximising what is known as the Evidence Lower BOund (ELBO) \citep{blei2}, which constitutes a bound on the likelihood of the data. The ELBO can be derived using the relationship in Eq. \ref{eq:elbo}:

\begin{equation}
    \log p_\theta(\mathbf{x}) = \mathbb{E}_{q(\mathbf{z}|\mathbf{x})} \left[\log \frac{p(\mathbf{x}|\mathbf{z}) p(\mathbf{z})}{q(\mathbf{z}|\mathbf{x})} + \log \frac{q(\mathbf{z}|\mathbf{x})}{p(\mathbf{z}|\mathbf{x})} \right]
    \label{eq:elbo}
\end{equation}

Here, the second term on the right hand side represents the Kullback-Liebler divergence between the approximating posterior $q(\mathbf{z}|\mathbf{x})$ and the true posterior $p(\mathbf{z}|\mathbf{x})$. As we do not have access to the true posterior, we specify a prior $p(\mathbf{z})$ as well as a family of (tractable) posterior distributions, and minimize the divergence between the prior and the posterior whilst also attempting to maximise the model evidence. The VAE provides the means to scale this amortized variational inference to intractable, high-dimensional problems, and minimizes the negative log likelihood over a dataset of $N$ samples by adjusting the parameters of neural networks $\{\theta, \phi\}$ according to the ELBO:

\begin{equation}
\begin{split}
\frac{1}{N} \sum_{i=1}^{n}-\log p_{\theta}(\mathbf{x}_i) \leq \\
\frac{1}{N} \sum_{i=1}^{N}\left(-\mathbb{E}_{q_{\phi}\left(\mathbf{z} | \mathbf{x}_i\right)}\left[\log p_{\theta}\left(\mathbf{x}_i | \mathbf{z}\right)\right]+ \beta\mathbb{D}_{KL}\left[q_{\phi}(\mathbf{z} | \mathbf{x}_i) \| p(\mathbf{z})\right]\right)
\end{split}
\label{eq:ELBO1}
\end{equation}

where $\beta = 1$ is used for the standard variational approximation procedure, but may be set empirically \citep{higgins}, annealed \citep{disentanglement} or optimized according to the Information Bottleneck principle \citep{alemi2017,tishby2015}. The first term in Eq.~\ref{eq:ELBO1} is the negative log-likelihood and is calculated in the form of a reconstruction error. The second term is the KLD between the approximating posterior and the prior, and therefore acts as a prior regularizer. Typically, the family of isotropic Gaussian distributions is chosen for the posterior $q_\phi(.)$, and an isotropic Gaussian with unit variance for the prior $p(\mathbf{z})$.

\textbf{Structure:} We seek to infer and disentangle the latent distribution into subsets of latent factors using VAEs. These latent subsets are $\{\mathbf{z}_t,\mathbf{z}_y, \mathbf{z}_c, \mathbf{z}_o\}$, which represent the instrumental factors on $t$, the risk factors on $y$, the confounders on both $t$ and $y$, and factors solely related to $\mathbf{x}$, respectively. Without inductive bias, consistently disentangling the latent variables into these factors would be impossible \citep{locatello} because there would be no guidance with which to assign specific information into specific latent factors. In TVAE this inductive bias is incorporated in a number of ways: firstly, by incorporating supervision and constraining $\mathbf{z}_t$ and $\mathbf{z}_y$ to be predictive of $t$ and $y$, respectively; secondly, by constraining $\mathbf{z}_c$ to be predictive of both $t$ and $y$; and finally, by employing diagonal-covariance priors (isotropic Gaussians) to encourage disentanglement and independence between latent variables.  The structural inductive bias on the model is such that $\mathbf{z}_y$, and $\mathbf{z}_t$, and $\mathbf{z}_c$ learn factors relevant to outcome and treatment, for which we provide explicit supervision, thereby leaving $\mathbf{z}_o$ for all remaining factors.

 \textbf{Identifiability:} In general, it is impossible to isolate the effect of $t \rightarrow y$ due to unobserved confounding \citep{Damou2019}, and this is why we make the assumption that the latent parents of the treatment and outcome may be inferred via noisy proxies present in the observed data. Under such assumptions, deep latent variable techniques have been shown to be able to infer hidden confounders from these proxy variables (see \textit{e.g.}, \citep{Allman2009, Lowe2020ACD, Montgomery2000, Louizos2017b, Parbhoo2020}). The usual assumption of ignorability then shifts from `all confounders are observed', to `all unobserved confounders have been inferred from proxies', both of which represent conditional exchangeability: $y(t) \indep t | \mathbf{z}$, such that the potential outcome is independent of the observed treatment given the inferred latent variables. We note that, empirically, variational models are not immune to convergence difficulties during optimization which may affect identifiability (see \textit{e.g.} \citep{Rissanen2021} for an exploration of the limitations of VAEs in the context of causal inference), and further work is required to establish bounds on the efficacy of these methods. 
 
   The proof for identifiability under the assumption of ignorability on the basis that relevant parent factors have been inferred from proxies and/or other observed variables, has been derived previously by \citep{Louizos2017b} and \citep{Zhang2020}. With reference to the graph depicted in Fig. \ref{fig:PGMs}(c), the factor $\mathbf{z}_o$ is $d$-separated from $t$ and $y$ given $\mathbf{z}$ (or $\mathbf{x}$), and does not affect the identification of the causal effect. \textit{i.e.}, the outcome under intervention $p(y|\mbox{do}(t),\mathbf{x})$ can be estimated from observational quantities given the inferred latent instrumental, risk, and confounding variables: $p(y|\mbox{do}(t),\mathbf{x}) = p(y|\mbox{do}(t),\mathbf{z}_{\{t,o,y,c\}})$. In turn, following the Markov property,  $p(y|\mbox{do}(t),\mathbf{z}_{\{t,o,y,c\}})= p(y|t,\mathbf{z}_y, \mathbf{z}_c)$ (see \citep{Zhang2020} and \citep{Louizos2017b} for additional information). 
   
    \textbf{Implementation:} We impose the priors and parameterizations denoted in Equations \ref{eq:paramsinf} and \ref{eq:paramsgen} , where $D_{(.)}$ is the number of dimensions in the respective variable (latent or otherwise), and $f_{1-11}$ and $h_{1-6}$ represent fully connected neural network functions.           Note that all Gaussian variance parameterizations are diagonal. In cases where prior knowledge dictates a discrete rather than continuous outcome, equivalent parameterizations to those in Eqs.~\ref{eq:paramsinf} and \ref{eq:paramsgen} may be employed. For example, in the IHDP dataset, the outcome data are standardized to have a variance of 1, and the outcome generation model becomes a Gaussian with variance also equal to 1. Note that separate treatment and outcome classifiers are used both during inference and generation ($\hat Q_q, \hat g_q$ and $\hat Q_p, \hat g_p$ resp.). The classifiers for inference have separate parameters to those use during generation. Predictors or classifiers of outcome incorporate the two-headed approach of \citep{Shalit2017}, and ground-truth $t$ are used for $\hat Q_q$ whereas generated samples $\hat{t}$ are used for $\hat Q_p$. For unseen test cases, either the ground-truth $t$ or an sampled treatment $\hat{t}$ from treatment classifier $\hat g_p$ may be used to simulate an outcome. During training $\hat{t}$ is used.

   \begin{equation}
\begin{gathered}
\mbox{\textbf{Inference:}} \\
q(\mathbf{z}_t|\mathbf{x}) = \prod_{d=1}^{D_{z_t}}\mathcal{N}(\mu_d = f_{1d}(\mathbf{x}), \sigma^2_d= f_{2d}(\mathbf{x})) \\  q(\mathbf{z}_y|\mathbf{x}) = \prod_{d=1}^{D_{z_y}}\mathcal{N}(\mu_d=f_{3d}(\mathbf{x}), \sigma^2_d=f_{4d}(\mathbf{x})) \\
q(\mathbf{z}_c|\mathbf{x}) =  \prod_{d=1}^{D_{z_c}}\mathcal{N}(\mu_d = f_{5d}(\mathbf{x}), \sigma^2_d = f_{6d}(\mathbf{x})) \\
q(\mathbf{z}_o | \mathbf{x}) = \prod_{d=1}^{D_{z_o}}\mathcal{N}(\mu_d = f_{7d}(\mathbf{x}), \sigma^2_d = f_{8d}(\mathbf{x}))\\
p(\hat{t}|\mathbf{z}_t, \mathbf{z}_c) = \mbox{Bern}(\hat g_q(.)) = \mbox{Bern}(f_9(\mathbf{z}_t, \mathbf{z}_c)) \\  p(\hat{y}|\mathbf{z}_y, \mathbf{z}_c, t) = \mbox{Bern}(\hat Q_q(.)) = \\ \mbox{Bern}(\hat{t} \cdot f_{10}(\mathbf{z}_y, \mathbf{z}_c) + (1 - \hat{t}) \cdot f_{11}(\mathbf{z}_y, \mathbf{z}_c)) \\ \\
\end{gathered}
\label{eq:paramsinf}
\end{equation}
\begin{equation}
\begin{gathered}
\mbox{\textbf{Generation:}} \\
p(\mathbf{z}_{\{o,t,c,y\}}) = \prod_d^{D_{\{z_{o,t,c,y}\}}} \mathcal{N}(z_{\{o,t,c,y\}d} | 0, 1) \\
 p(\hat{t}|\mathbf{z}_t, \mathbf{z}_c)= \mbox{Bern}(g_p(.)) = \mbox{Bern}( h_1(\mathbf{z}_t, \mathbf{z}_c))\\
p(\hat{y}|\mathbf{z}_y, \mathbf{z}_c, \hat{t}) = \mbox{Bern}(\hat Q_p(.)) = \\ \mbox{Bern}(\hat{t} \cdot h_2(\mathbf{z}_y, \mathbf{z}_c) + (1 - \hat{t}) \cdot h_3(\mathbf{z}_y, \mathbf{z}_c))\\
p(\hat{\mathbf{x}}_{bin} | \mathbf{z}_c, \mathbf{z}_o, \mathbf{z}_t, \mathbf{z}_y) = \mbox{Bern}(h_6(\mathbf{z}_c, \mathbf{z}_o, \mathbf{z}_t, \mathbf{z}_y))\\
p(\hat{\mathbf{x}}_{cont} | \mathbf{z}_c, \mathbf{z}_o, \mathbf{z}_t, \mathbf{z}_y) = \\ \prod_{d=1}^{D_{x_{cont}}}\mathcal{N}(x_{cont,d}|\mu_d = h_4(\mathbf{z}_{\{c,o,t,y \}}), \sigma^2_d = h_5(\mathbf{z}_{\{c,o,t,y \}}))
\end{gathered}
\label{eq:paramsgen}
\end{equation}

The parameters for these neural networks are learnt via variational Bayesian approximate inference \citep{kingma} according to the objective in Eq. \ref{eq:ted}:
 \begin{equation}
\begin{aligned} \mathcal{L}_i^{\mathrm{ELBO}}= \sum_i^N \mathbb{E}_{q_{c} q_{t} q_{y}q_{o}}\left[\log p\left(\hat{\mathbf{x}}_i | \mathbf{z}_{t}, \mathbf{z}_{c}, \mathbf{z}_{y},\mathbf{z}_{o}\right) \right. \\ + \left.\log p\left(\hat{t}_i | \mathbf{z}_{t}, \mathbf{z}_{c}\right) + \log p\left(\hat{y}_i | t_i, \mathbf{z}_{y}, \mathbf{z}_{c}\right)\right] & \\
- \left[ D_{K L}\left(q\left(\mathbf{z}_{t} | \mathbf{x}_i \right)|| p\left(\mathbf{z}_{t}\right)\right) +D_{K L}\left(q\left(\mathbf{z}_{c} | \mathbf{x},\right) \|p\left(\mathbf{z}_{c}\right)\right) \right. &\\ \left.
+D_{K L}\left(q\left(\mathbf{z}_{y} | \mathbf{x}_i\right) \|p\left(\mathbf{z}_{y}\right)\right)
+D_{K L}\left(q\left(\mathbf{z}_{o} |\mathbf{x}_i\right) \|p\left(\mathbf{z}_{o}\right)\right)\right]
\end{aligned}
\label{eq:ted}
\end{equation}

 \textbf{Targeted Regularization:} We now introduce the targeted regularization, the purpose of which is to encourage the outcome to be independent of the treatment assignment. Following Eq.~\ref{eq:update}, we define the fluctuation sub-model and corresponding logistic loss for estimating $\epsilon$ in Eqs. \ref{eq:update2} and \ref{eq:update3}: 
 
 \begin{equation}
\begin{split}
    \hat{Q}^1(\hat g, t_i,\mathbf{z}^y_i,\mathbf{z}^c_i, \hat \epsilon) = \sigma \left[ \sigma^{-1}[ \hat Q^0( t_i,\mathbf{z}^y_i,\mathbf{z}^c_i)] \right.  \\ \left.  + \hat \epsilon \left( \frac{I(t_i=1)}{\hat g(t_i=1; \mathbf{z}^t_i,\mathbf{z}^c_i)} - \frac{I(t_i=0)}{\hat g(t_i=0; \mathbf{z}^t_i,\mathbf{z}^c_i)}\right) \right]
    \end{split}
    \label{eq:update2}
\end{equation}

\begin{equation}
\begin{split}
\xi_i(\hat{Q}^1; \hat \epsilon) = - y_i \log(\hat{Q}^1(\hat g, t_i,\mathbf{z}^y_i,\mathbf{z}^c_i, \hat \epsilon)) \\ - (1-y_i)\log(1-\hat{Q}^1(\hat g, t_i,\mathbf{z}^y_i,\mathbf{z}^c_i, \hat \epsilon))
\end{split}
\label{eq:update3}
\end{equation}
 
 In Eq.~\ref{eq:update2}, $I$ is the indicator function. For an unbounded regression loss, mean squared error loss may be used. Note that the logistic loss is suitable for continuous outcomes bounded between 0 and 1 (see \cite[pp.121:132]{vanderLaan2011} for proof). Putting it all together, we then optimize to find generative parameters for functions $h_{1-6}$, inference parameters for functions $f_{1-12}$, and estimated fluctuation parameter $\hat \epsilon$ in Eq. \ref{eq:TVAEobj}:

\begin{equation}
\begin{split}
    \mathcal{L} = \min \biggl[ \sum_i^N \biggl( \mathcal{L}^{\mathrm{ELBO}}_i + \lambda_{TL}\xi_i(\hat Q, \hat g, \hat \epsilon) \biggr) \biggr]; \\ \left.\frac{\partial}{\partial \epsilon} \mathcal{L}^*\right\vert_{\epsilon=0} = \bar{IC}^* = 0
    \end{split}
    \label{eq:TVAEobj}
\end{equation}\\

 In Eq. \ref{eq:TVAEobj}, $\lambda_{TL}$ represents a hyperparameter for the targeted regularization weight. At convergence $\hat \epsilon=0$ and $\hat Q$ and $\hat g$ become consistent estimators, thereby satisfying the conditions for the EIC (see Eq.~\ref{eq:EIC} and reference \cite[pp125:128]{vanderLaan2011}).
 
 One aspect of TVAE that bears mentioning (and which differentiates TVAE from another recent contribution \citep{Shi2019} that uses targeted regularization) is that the gradients resulting from $\xi$ are \textit{not} taken with respect to the propensity score arms $\hat g_p$ or $\hat g_q$. Targeted learning is concerned with debiasing the outcome classifier $\hat Q$ using propensity scores from $\hat g$. In other words, assuming the propensity scores are consistently estimated, the targeted learning regularizer is intended to affect the outcome classifier only, and \textit{not} the propensity score estimator. It is therefore more theoretically aligned (with the targeted learning literature) to apply regularization to the outcome estimator $\hat Q$, and not to $\hat g$. As per Eq. \ref{eq:update}, in TMLE, $\hat g$ is assumed to be a consistent estimator, forming part of the debiasing update process for $\hat Q$, but it is not subject to update itself. In order to prevent the regularization from affecting the propensity arms, the gradients from the regularizer are only taken with respect to all parameters that influence this outcome classifier (which include upstream parameters for $\hat Q_q$ as well as the more direct parameters $\hat Q_p$). We use Pytorch's `detach' method on the propensity scores when calculating the targeted regularization. This method decouples the propensity score arm from backpropagation relating to the computation of the regularization value.

   The notable aspects of our model are as follows: the introduction of a new latent variable $\mathbf{z}_o$ for factors unrelated to outcome and/or treatment to aid the recovery of the true underlying structure and, as far as we are aware, the first incorporation of targeted learning in a deep latent variable approach.

\section{Related Work}
There are a number of ways to mitigate the problems associated with the confounding between the covariates and the treatment. For a review on such methods, readers are pointed to the recent surveys by \citep{Yao2020, Guo2020}. Here we consider methods that utilize neural networks as part of their models, but note that many non-neural network methods exist \citep{Chernozhukov2017, vanderLaan2011, Rubin2005, Hill2011, Athey2016}.

Perhaps the most similar works to ours are those of Dragonnet \citep{Shi2019}, TEDVAE \citep{Zhang2020}, and Intact-VAE \citep{Wu2021}. We discuss the differences between these and TVAE in turn. Dragonnet is a non-generative model which incorporates the same targeted learning regularization process which allows for the simultaneous optimization of $Q$ and $\epsilon$. However, the method sacrifices the ability to estimate conditional treatment effects to achieve good estimation of the average treatment effect across the sample. Indeed, they do not report PEHE. Finally, Dragonnet applies regularization to the entire network, whereas we `target' the regularization to the outcome prediction arm by restricting the backpropagation of gradients.

TEDVAE is one of the few variational generative models for causal inference which builds on CEVAE \citep{Louizos2017b} and seeks disentanglement of the latent instrumental, risk, and confounding factors. However, it has no means to allocate latent variables that are unrelated to treatment and/or outcome. The advantage of including factors $\mathbf{z}_o$ with a variational penalty is that the model has the option to use them, or not to use them, depending on whether they are necessary (\textit{i.e.}, KL is pushed to zero). It is important not to force factors unrelated to treatment and outcome into $\mathbf{z}_{\{c,y,t \}}$ because doing so restricts the overlap between the class of models that can be represented using TEDVAE, and the class of models describing the true process.

Intact-VAE seeks to address some of the issues associated with the use of deep latent variable models and causal identification, in particular when the method's performance hinges on its ability to recover the posterior distribution of unobserved confounders. They do not disentangle risk and instrumental variables from confounders, focusing instead on the successful recovery of the confounder via the use of a novel development of the prognostic score which they refer to as $B^*$-scores.

Other methods include GANITE \citep{Yoon2018} which requires adversarial training, and may therefore be more difficult to optimise. PM \citep{Schwab2019}, SITE \citep{Yao2018}, and MultiMBNN \citep{Sharma2020} incorporate propensity score matching. TARNET \citep{Shalit2017} inspired the two-headed outcome arm in our TVAE, as well as the three-headed architecture in \citep{Shi2019}. RSB \citep{Zhang2019} incorporates regularization based on the Pearson Correlation, intended to reduce the association between latent variables predictive of treatment assignment and those predictive of outcome.

\section{Experiments}
We begin by performing an ablation study on our synthetic TVAESynth dataset, comparing (a) TVAE (base) which is equivalent to TEDVAE (b) TVAE with $\mathbf{z}_o$, and (c) TVAE with both with $\mathbf{z}_o$ and targeted regularization $\xi$ during training. In order to evaluate the benefits of introducing $\mathbf{z}_o$, we ensure that the total number of latent dimensions remains constant. 

We then utilize 100 replications of the semi-synthetic \emph{Infant Health and Development Program (IHDP)} dataset \citep{Hill2011, Gross1993}\footnote{Available from \url{https://www.fredjo.com/}} The linked version (footnote) corresponds with usual setting A of the NPCI data generating package \citep{Dorie2016} (see \citep{Shi2019, Shalit2017, Yao2018}) and comprises 608 untreated and 139 treated samples (747 in total). There are 25 covariates, 19 of which are discrete/binary, and the rest are continuous. The outcome for the IHDP data is continuous and unbounded. Similarly to \citep{Louizos2017b, Shalit2017} and others, we utilize a 60/30/10 train/validation/test split. We evaluate our network on the Average Treatment Effect estimation error (eATE), and the Precision in Estimation of Heterogeneous Effect (PEHE).

We also utilize the job outcomes dataset (\emph{Jobs}) \citep{Lalonde1986, Smith2005}.\footnote{Available from \url{https://users.nber.org/~rdehejia/data/.nswdata2.html}.} Unlike the IHDP dataset, Jobs is real-world data with a binary outcome. We follow a similar procedure to \citep{Shalit2017} who indicate that they used the Dehejia-Wahba \citep{Dehejia2002} and PSID comparison sample. This sample comprises 260 treated samples and 185 control samples, along with the PSID comparison group comprising 2490 samples'. The dataset contains a mixture of observational and RCT data. Similarly to \citep{Louizos2017b, Shalit2017} and others, we utilize a 56/24/20 train/validation/test split, and undertake 100 runs with varying random split allocations in order to acquire an estimate of average performance and standard error. Note that, between models, the same random seed is used both for initialization as well as dataset splitting, and therefore the variance due to these factors is equivalent across experiments. As per \citep{Louizos2017b, Shalit2017, Yao2018}, for the Jobs dataset (for which we have only partial effect supervision) we evaluate our network on the Average Treatment effect on the Treated error:

\begin{equation}
\begin{split}
eATT = ||T_1|^{-1}\sum_{i\in T_1}y_i - |T_0|^{-1}\sum_{j\in T_0}y_j \\ - |T_1|^{-1}\sum_{i\in T_1}(\hat Q(1,\mathbf{x}_i) - \hat Q(0,\mathbf{x}_i))|
\end{split}
\label{eq:ATT}
\end{equation}

where $T = T_1 \cup T_0$ constitutes all individuals in the RCT, and the subscripts denote whether or not those individuals were in the treatment (subscript 1) or control groups (subscript 0). The first two terms in Eq.~\ref{eq:ATT} comprise the true ATT, and the third term the estimated ATT. We may use the policy risk as a proxy for PEHE:

\begin{equation}
\begin{split}
\mathcal{R}_{pol} = 1 - \left( \mathbb{E}[y(1)|\pi(\mathbf{x}) = 1 ]p(\pi(\mathbf{x})=1) \right. \\  \left. + \mathbb{E}[y(0)|\pi(\mathbf{x}) = 0]p(\pi(\mathbf{x}) =0 ) \right)
\end{split}
\end{equation}

where $\pi(\mathbf{x}_i) = 1$ is the policy to treat when $\hat y_i(1) - \hat y_i(0) > \alpha$, and $\pi(\mathbf{x}_i) = 0$ is the policy not to treat otherwise \citep{Yao2018, Shalit2017}. $\alpha$ is a treatment threshold. This threshold can be varied to understand how treatment inclusion rates affect the policy risk. We set $\alpha=0$, as per \citep{Shalit2017, Louizos2017b}.

\begin{figure}[h!]
\centering
\includegraphics[width=0.6\linewidth]{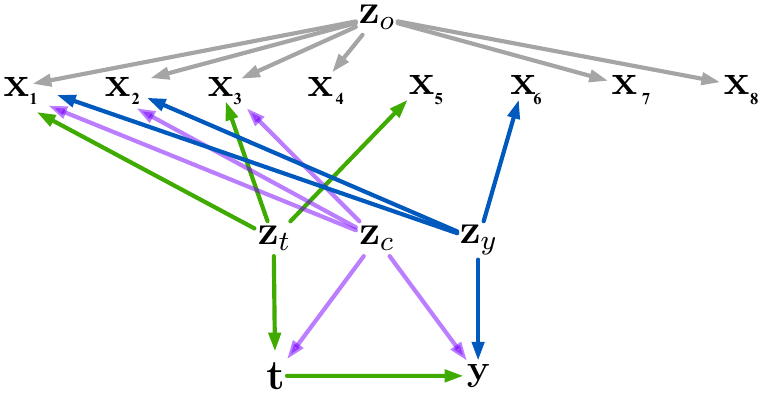}
\caption{The DAG for TVAESynth dataset.}
\label{fig:TVAESynth}
\end{figure}

Finally, we introduce a new synthetic dataset named \emph{TVAESynth} which follows the structure shown in Figure \ref{fig:TVAESynth}, and relationships in Equations \ref{eq:tvaesnthfirst}-\ref{eq:tvaesnthend}. While the weightings are chosen relatively arbitrarily, the structure is intentionally designed such that there are a mixture of exogenous and endogenous covariates. This enables us to compare the performance of TVAE with and without $\mathbf{z}_o$ (keeping the total number of latent dimensions constant). The dataset was designed such that not all covariates are exogenous and so that there exist some latent factors unrelated to outcome and treatment. Thus, we should expect an improvement in performance to occur with the introduction of $\mathbf{z}_o$, demonstrating the importance of incorporating inductive bias that closely matches the true structure. Note that while these datasets vary in whether the outcome variable is continuous (IHDP, TVAESynth) or binary (Jobs), the treatment variable is always binary. We leave an application to data with continuous treatment effects to future work.

\begin{equation}
\begin{split}
    \mathbf{U}_{z_o, z_c, z_t, z_y, y} \sim \mathcal{N}(\mathbf{0,1}) \; \; \; \\
    \mathbf{U}_{x_1, x_4, t} \sim \mbox{Bernoulli}(0.5) \; \; \; \mathbf{U}_{x_{2:3}, x_{5:8}} \sim \mathcal{N}(\mathbf{0,1})\\
        \mathbf{z}_o = \mathbf{U}_{z_o} \; \; \; \mathbf{z}_y = \mathbf{U}_{z_y} \; \; \; \mathbf{z}_t = \mathbf{U}_{z_t} \; \; \; \mathbf{z}_c = \mathbf{U}_{z_c}
\end{split}
\label{eq:tvaesnthfirst}
\end{equation}

\begin{equation}
\begin{split}
\mathbf{x}_1 \sim \mbox{Bernoulli}(\sigma(\mathbf{z}_t + 0.1(\mathbf{U}_{x_1}-0.5))) \; \; \; \\
\mathbf{x}_2 \sim \mathcal{N}(0.4\mathbf{z}_o + 0.3\mathbf{z}_c + 0.5 \mathbf{z}_y + 0.1\mathbf{U}_{x_2}, 0.2)\\
\mathbf{x}_3 \sim \mathcal{N}(0.2\mathbf{z}_o + 0.2\mathbf{z}_c + 1.2 \mathbf{z}_t + 0.1\mathbf{U}_{x_3}, 0.2)\\
\mathbf{x}_4 \sim \mbox{Bernoulli}(\sigma(0.6\mathbf{z}_o + 0.1(\mathbf{U}_{x_4}-0.5)))\\
\mathbf{x}_5 \sim \mathcal{N}(0.6\mathbf{z}_t + 0.1\mathbf{U}_{x_5}, 0.1)\\
\mathbf{x}_6 \sim \mathcal{N}(0.9\mathbf{z}_y + 0.1\mathbf{U}_{x_6}, 0.1)\\
\mathbf{x}_7 \sim \mathcal{N}(0.5\mathbf{z}_o + 0.1\mathbf{U}_{x_7}, 0.1)\\
\mathbf{x}_8 \sim \mathcal{N}(0.5\mathbf{z}_o + 0.1\mathbf{U}_{x_8}, 0.1)
\end{split}
\end{equation}

\begin{equation}
t_p = \sigma(0.2\mathbf{z}_c + 0.8\mathbf{z}_t + 0.1\mathbf{U}_t) \; \; \; t \sim \mbox{Bernoulli}(t_p)
\end{equation}

\begin{equation}
y := 0.2\mathbf{z}_c + 0.5\mathbf{z}_y \* t + 0.2t +  0.1\mathbf{U_y}
\label{eq:tvaesnthend}
\end{equation}

  When estimating treatment effects, 100 samples are drawn for each set of input covariates. We provide results for both within sample and out-of-sample performance. Note that within sample and out-of-sample results are equally valid for treatment effect estimation because the network is never supervised on treatment effect \citep{Shi2019}.

\subsection{Architecture and Hyperparameters}

\begin{figure*}
\centering
\includegraphics[width=1\linewidth]{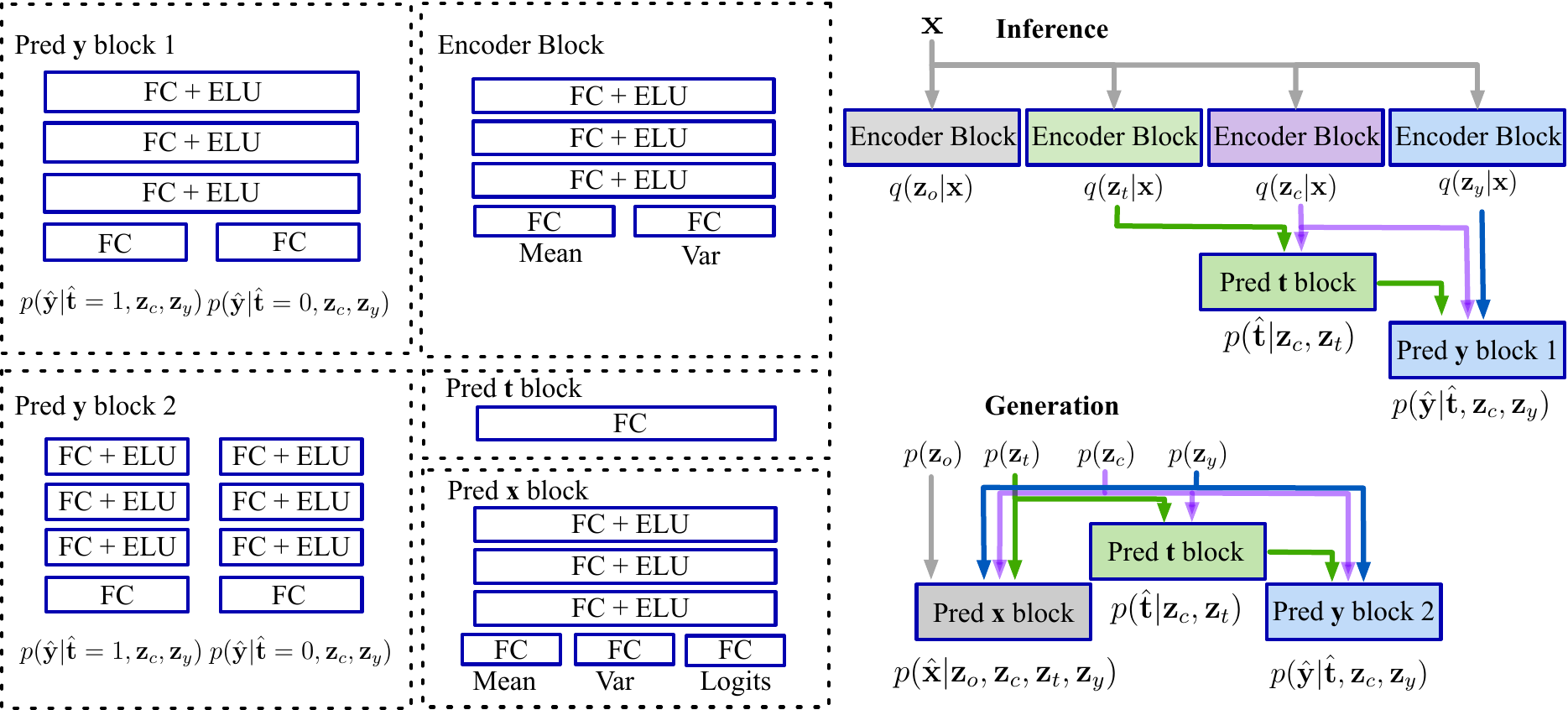}
\caption{Block architectural diagram for TVAE's inference and generation models (number of layers in Pred $y$ 1 \& 2 and Pred $\mathbf{x}$ blocks varies by experiment, as does the number of neurons in each layer (see details in text).}
\label{fig:archi}
\end{figure*}

The architecture is shown in Fig.~\ref{fig:archi}. For continuous outcomes we standardize the values and model as a Gaussian with a fixed variance of 1, and a mean determined by the outcome arm. All binary variables in the model are modelled as Bernoulli distributed with a probability determined by the associated neural network function. 

Hyperparameters\footnote{Bold font indicates the settings that were finally used.} for IHDP experiments were: hidden layers: $\mathbf{3}$; the weight on targeted regularization $\lambda_{TL} = \{0.0, 0.1, 0.2, \mathbf{0.4}, 0.6, 0.8, 1.0\}$; learning rate $LR =$ \{1e-3, 1e-4, \textbf{5e-5}\}; hidden neurons $=$\textbf{300}; layers $=\mathbf{4}$; dimensionality of latent factors was $D_{z_t} = D_{z_t} = \mathbf{10}, D_{z_c} = \mathbf{15}, D_{z_o} = \mathbf{5}$; batch size of $\mathbf{200}$; epochs $\mathbf{200}$; weight regularization \textbf{1e-4}; and learning rate decay \textbf{5e-4}. Hyperparameters for Jobs experiments were: hidden layers: $\mathbf{3}$; the weight on targeted regularization $\lambda_{TL} =$ \{0.0, \textbf{0.1}, 0.2, 0.4, 0.6, 0.8, 1.0\}; learning rate $LR =$ \{5e-5, \textbf{1e-5}\}; hidden neurons $=$\textbf{200};  layers $=\mathbf{2}$; dimensionality of latent factors was $D_{z_t} = D_{z_t} = \mathbf{6}, D_{z_c} = \mathbf{8}, D_{z_o} = \mathbf{4}$; batch size of $\mathbf{200}$; epochs $\mathbf{200}$; weight regularization \textbf{1e-4}; and learning rate decay \textbf{5e-4=3}. Hyperparameters for TVAESynth experiments were: hidden layers: $\mathbf{2}$; the weight on targeted regularization $\lambda_{TL} =$ \{0.0, \textbf{0.1}, 0.2, 0.4, 0.6, 0.8, 1.0\}; learning rate $LR =$ \textbf{5e-5}; hidden neurons $=$\textbf{20};  layers $=\mathbf{2}$; dimensionality of latent factors was $D_{z_t} = D_{z_t} = D_{z_c} = \mathbf{2}, D_{z_o} = \mathbf{1}$; batch size of $\mathbf{200}$; epochs $\mathbf{40}$; weight regularization \textbf{1e-4}; and learning rate decay \textbf{5e-3}. All models were optimized using Adam \citep{adam}.

For model selection we use the minimum validation loss on the total objective function \citep{Louizos2017b, Zhang2020}. Whilst some model selection heuristics exist that serve as surrogates for the eATE itself (\textit{e.g.}, see \citep{Hassanpour2019, Athey2016}) we take the same view as \citep{Zhang2020}, in that the development of our model `should be self-sufficient and not rely on others'. For all experiments, we undertake 100 replications and provide mean and standard error. There may be room to improve performance with further tuning. However, given that the tuning of hyperparameters in a causal inference paradigm is problematic in general, we intentionally limited the search space. 

The network is coded using Pyro \citep{Bingham2019pyro} and is an extension of the code by \citep{Zhang2020} available here: \url{https://github.com/WeijiaZhang24/TEDVAE}. We train on a GPU (\textit{e.g.}, NVIDIA 2080Ti) driven by a 3.6GHz Intel I9-9900K CPU running Ubuntu 18.04. Training 200 epochs of the IHDP dataset (training split of 450 samples) takes approx. 35 seconds (0.175s per epoch). 


\begin{table*}[ht!]
\centering
\caption{Means and standard errors for the ablation study using TVAESynth. `oos' is out-of-sample, `ws' is within sample. `$+\mathbf{z}_o$' indicates the introduction of the miscellaneous factors, `$+ \mathbf{z}_o^*$' indicates the introduction of miscellaneous factors but \textit{without} changing the dimensionality of $\mathbf{z}_c$ thereby increasing total latent capacity, `$+\xi$' indicates our targeted regularization (with selected backpropagation), `$+\xi^*$' indicates targeted reg. equivalent to the one used by \citep{Shi2019} (\textit{i.e.}, with gradients applied to all upstream parameters), the $\xi$ subscript indicates its weight in the loss.}
\begin{tabular}{lllll}
\\
\textbf{Method} & $\sqrt{\epsilon_{PEHE}}$ ws & $\sqrt{\epsilon_{PEHE}}$ oos & $\epsilon_{ATE}$ ws &$\epsilon_{ATE}$ oos \\ \hline
TVAE (base/TEDVAE) & .179$\pm$.003 & .178$\pm$.003 & .128$\pm$.005 & .128$\pm$.005 \\
TVAE + $\mathbf{z}_{o}^*$ &  .174$\pm$.003 & .173$\pm$.003 & .121$\pm$.005 & .120$\pm$.005\\
TVAE + $\mathbf{z}_o$ &  .166$\pm$.003 & .166$\pm$.003 & .069$\pm$.004 & .069$\pm$.004\\
TVAE + $\xi_{\lambda=0.1}$ &  .171$\pm$.003 & .170$\pm$.003 & .122$\pm$.004 & .121$\pm$.004\\
TVAE + $\mathbf{z}_o$ + $\xi_{\lambda=0.1}^*$  & \bftab .151$\pm$\bftab .002 & \bftab .150$\pm$\bftab.003 & \bftab.048$\pm$\bftab.003 & \bftab .048$\pm$\bftab.003 \\
TVAE + $\mathbf{z}_o$ + $\xi_{\lambda=0.1}$ (full model) & \bftab .150$\pm$\bftab .002 & \bftab .150$\pm$\bftab.003 & \bftab.048$\pm$\bftab.003 & \bftab .048$\pm$\bftab.003 \\
\hline
\end{tabular}
\label{tab:ablation}
\end{table*}

\subsection{Results}
\textbf{Ablation Study Results} are shown in Table \ref{tab:ablation}. They demonstrate that both eATE and PEHE are significantly improved by the incorporation of $\mathbf{z}_o$ or targeted regularization, with a combination of the two yielding the best results for both within sample and out of sample testing. The fact that TVAE $+\mathbf{z}_o$ outperforms TVAE $+\mathbf{z}_o^*$ despite the latter having a larger latent capacity, suggests that reducing the capacity of the latent space has a beneficial, regularizing effect. Based on the results of this ablation, the benefits of this regularizing effect appear to be distinct from the benefits that derive from the addition of miscellaneous factors. Finally, the results indicate negligible empirical benefits to restricting the backgpropagation of the regularizer to non-propensity related parameters. However, our restriction of the backpropagation more closely aligns with the original TMLE and efficient influence curve theory, and we therefore retain this feature for the remaining experiments. 

\textbf{IHDP Results} are shown in Table \ref{tab:IHDPresults} and indicate state of the art performance for both within sample and out-of-sample eATE and PEHE. They corroborate the ablation results in Table \ref{tab:ablation}, in that the incorporation of $\mathbf{z}_o$ and targeted regularization result in monotonic improvement above TEDVAE. TVAE is outperformed only by Dragonnet on within-sample eATE performance. However, this method does not provide estimations for individual CATE, and is limited to the estimation of average treatment effects.


\textbf{Jobs Results} are shown in Table \ref{tab:JOBSresults}. GANITE was found to perform the best across most metrics, although this method has been argued to be more reliant on larger sample sizes than others, on the basis that it performs relatively poorly on the smaller IHDP dataset \citep{Yoon2018}. Furthermore, GANITE relies on potentially unstable/unreliable adversarial training \citep{moyer1, lezama, gabbay2019}. Finally, TVAE outperforms GANITE on eATT, is consistent (beyond 2 d.p.) across out-of-sample and within-sample evaluations and has a lower standard err, and is competitive across all metrics. On this dataset, the concomitant improvements associated with the additional latent factors and targeted learning were negligible.
\begin{table*}[ht!]
\centering
\caption{Means and standard errors for evaluation on IHDP \citep{Hill2011}. Results from: \citep{Louizos2017b, Shalit2017,Zhang2020, Yoon2018}. `oos' is out-of-sample and `ws' is within sample. `+ $\mathbf{z}_o$' indicates the introduction of the miscellaneous factors, and `+ $\xi$' indicates targeted reg. with subscript indicating its weight in the loss.}
\begin{tabular}{lllll} 
\\
\textbf{Method} & $\sqrt{\epsilon_{PEHE}}$ ws & $\sqrt{\epsilon_{PEHE}}$ oos & $\epsilon_{ATE}$ ws &$\epsilon_{ATE}$ oos \\ \hline
TMLE \citep{vanderLaan2018}   & 5.0$\pm$.20  &  - & .30$\pm$.01  &  - \\
CEVAE \citep{Louizos2017b}       & 2.7$\pm$.10  & 2.6$\pm$.10  & .34$\pm$.01  & .46$\pm$.02  \\
TARNet \citep{Shalit2017}       & .88$\pm$.00  & .95$\pm$.00  & .26$\pm$.01  & .28$\pm$.01  \\
CFR-MMD \citep{Shalit2017}      &  .73$\pm$.00 & .78$\pm$.00  & .30$\pm$.01  & .31$\pm$.01  \\
CFR-Wass \citep{Shalit2017}     &  .71$\pm$.00 & .76$\pm$.00  & .25$\pm$.01  & .27$\pm$.01 \\
TEDVAE \citep{Zhang2020} & .62$\pm$.11 &  .63$\pm$.12 &- & .20$\pm$.05  \\
IntactVAE \citep{Wu2021} & .97$\pm$.04 & 1.0$\pm$.05  & .17$\pm$.01&  .21$\pm$.01 \\
GANITE \citep{Yoon2018} & 1.9$\pm$.40 & 2.4$\pm$.40 & .43$\pm$.05 & .49$\pm$.05 \\
Dragonnet w/ t-reg \citep{Shi2019} & - & - & \bftab.14$\pm$\bftab.01 & .20$\pm$.01 \\
TVAE ($\mathbf{z}_0$, $\xi_{\lambda=0.0}$) & \bftab.57$\pm$\bftab.03 & \bftab.57$\pm$\bftab.03 & \bftab.16$\pm$\bftab.01 & \bftab.16$\pm$\bftab.01 \\
TVAE ($\mathbf{z}_0$, $\xi_{\lambda=0.4}$) & \bftab.52$\pm$\bftab.02 & \bftab.54$\pm$.02 & \bftab.15$\pm$.01 & \bftab.16$\pm$.01 \\
\hline
\end{tabular}
\label{tab:IHDPresults}
\end{table*}

\begin{table}[h!]
\centering
\caption{Means and standard errors for evaluation on Jobs. Results taken from: \citep{Louizos2017b, Zhang2020}. `oos' is out-of-sample and `ws' is sample. `+ $\mathbf{z}_o$' indicates the introduction of the miscellaneous factors, and `+ $\xi$' indicates targeted reg. with subscript indicating its weight in the loss.}
\begin{tabular}{lllll} 
\\ 
\textbf{Method} & $R_{pol}$ ws & $R_{pol}$ oos & $\epsilon_{ATT}$ ws &$\epsilon_{ATT}$ oos \\ \hline
   TMLE \citep{vanderLaan2018}   &  .22$\pm$.00 &  - & .02$\pm$.01  & - \\
            CEVAE \citep{Louizos2017b}    &  .15$\pm$.00 & .26$\pm$.00  & .02$\pm$.01  & .03$\pm$.01 \\
               TARNet  \citep{Shalit2017}    & .17$\pm$.00  & .21$\pm$.00  & .05$\pm$.02  & .11$\pm$.04 \\
                  CFR-MMD  \citep{Shalit2017}   & .18$\pm$.00  & .21$\pm$.00  & .04$\pm$.01  & .08$\pm$.03 \\
                     CFR-Wass  \citep{Shalit2017}   &  .17$\pm$.00 & .21$\pm$.00  & .04$\pm$.01  & .09$\pm$.03 \\
                     GANITE \citep{Yoon2018} & \bftab.13$\pm$.00 & \bftab.14$\pm$.00 & \bftab.01$\pm$.01 & .06$\pm$.03\\
                     TEDVAE \citep{Zhang2020} & - & -&.06$\pm$.00 & .06$\pm$.00 \\
                    TVAE ($\mathbf{z}_0$, $\xi_{\lambda=0}$) & .16$\pm$.00 & .16$\pm$.00  & \bftab.01$\pm$.00 &  \bftab.01$\pm$.00 \\
                    TVAE ($\mathbf{z}_0$, $\xi_{\lambda=1}$) & .16$\pm$.00 & .16$\pm$.00 &  \bftab.01$\pm$.00 & \bftab.01$\pm$.00 \\
                  \hline
\end{tabular}
\label{tab:JOBSresults}
\end{table}

\section{Conclusion}

We aimed to improve existing latent variable models for causal parameter estimation in two ways: Firstly, by introducing a latent variable to model factors unrelated to treatment and outcome, thereby enabling the model to more closely reflect the data structure; and secondly, by incorporating a targeted learning regularizer with selected backpropagation to further debias outcome predictions. Our experiments demonstrated concomitant improvements in performance, and our comparison against other methods demonstrated TVAE's ability to compete with and/or exceed state of the art for both individual as well as average treatment effect estimation. In spite of TVAE's promising performance, it is worth remembering the method's limitations. Firstly, we assume that there are sufficient proxy variables present in the observations to facilitate inference of the latent factors. Secondly, variational approaches are approximate and their performance depends on a number of aspects, such as the choice of posterior distribution, and on optimization convergence. The latter point can significantly affect identifiability of causal effects \citep{Rissanen2021}. 

For future work, we plan apply TVAE to longitudinal data with continuous or categorical treatment, to explore the use of TVAE in inferring hidden confounders from proxies in the dataset, and to explore the bounds on identifiability associated with the use of structured models in combination with targeted regularization. Additionally, it was noted from the ablation study results that the restriction of regularization gradients did not yield a significant change in performance when compared with applying the regularization to the entire network. It would also be highly valuable to establish to what extent the properties of targeted approaches (\textit{e.g.}, double robustness) carry over to neural network estimators which use targeted regularization. Finally, the ablation results indicated that part of the improvement associated with the introduction of $\mathbf{z}_o$ is associated with a regularizing effect relating to the reduction in the dimensionality of $\mathbf{z}_c$. This aspect of the model's behavior also deserves investigating.

\bibliography{NN.bib}
\bibliographystyle{iclr2021_conference}

\end{document}